\documentclass[fleqn,10pt]{wlscirep}
\usepackage[utf8]{inputenc}
\usepackage[T1]{fontenc}
\usepackage{caption}
\usepackage{subcaption}
\usepackage{wrapfig}
\title{ELRA: Exponential learning rate adaption gradient descent optimization method}

\author[1,*]{Alexander Kleinsorge}
\author[1]{Stefan Kupper}
\author[1]{Alexander Fauck}
\author[1]{Felix Rothe}
\affil[1]{Technical University Wildau, Telematics, Wildau, 15745, Germany}

\affil[*]{alexander.kleinsorge@th-wildau.de} 

\begin{abstract}
We present a novel, fast (exponential rate adaption), ab initio (hyper-parameter-free) gradient based optimizer algorithm. The main idea of the method is to adapt the learning rate $ \alpha $ by situational awareness, mainly striving for orthogonal neighboring gradients. The method has a high success and fast convergence rate and does not rely on hand-tuned parameters giving it greater universality. It can be applied to problems of any dimensions $n$ and scales only linearly (of order $O(n)$) with the dimension of the problem. It optimizes convex and non-convex continuous landscapes providing some kind of gradient. In contrast to the Ada-family (AdaGrad, AdaMax, AdaDelta, Adam, etc.) the method is rotation invariant: optimization path and performance are independent of coordinate choices. The impressive performance is demonstrated by extensive experiments on the MNIST benchmark data-set against state-of-the-art optimizers.
We name this new class of optimizers after its core idea \textbf{E}xponential \textbf{L}earning \textbf{R}ate \textbf{A}daption -- \textbf{ELRA}. We present it in two variants c2min and p2min with slightly different control.
The authors strongly believe that ELRA will open a completely new research direction for gradient descent optimizers.
\end{abstract}
\begin{document}

\flushbottom
\maketitle
%
%
\thispagestyle{empty}


\section*{Introduction}
Numerical optimization of functions obviously relies on information obtained from the function $f(x)$ landscape. One key problem is that usually we are lacking meaningful global information about $f(x)$ making it necessary to rely instead on local information. Approaches using this local information range from using the function value in physics-inspired relaxation approaches \cite{coolmomentum} to algorithms using directly the topographical structure of the function landscape such as gradient descent-like approaches to biology inspired algorithms such as swarm optimization \cite{psocitation}.\\
Among these, the gradient descent-like methods have the longest history and are in high dimensional problems (e.g.\ DNN) the only practically applicable algorithms (due to their linear scaling with the problems dimension). In these approaches the gradient $G=\nabla f(x)$ of the function $f(x)$ is computed and thus also the best descent direction $-G$. However, while the idea of going downhill is obviously reasonable in the first place, no information whatsoever can be obtained about how long an optimal step should be, e.g.\ the optimal step-length $\lambda(x)=||\alpha \cdot \nabla(f(x))||$ is unknown. The parameter $\alpha$ is referred to as the step size or learning rate. Most current gradient based algorithms use a learning rate $\alpha$ independent of $x$ (but sometimes dependent on the time or step count $t$), estimated by some initial try and error test runs. This holds in particular for the Ada-family\footnote{Such as: AdaGrad, RMSProp, AdaDelta, Adam, etc., they all scale the gradient components individually (precondition-like)} of optimizers, widely used for training neural networks. To eliminate the initial tuning of $\alpha$, there are some modern approaches which adapt the learning rate $\alpha$ such as AdaDelta or the algorithms described in Prodigy2023\cite{mishchenko2023prodigy} or
DogSgd2023\cite{ivgi2023dog}. Yet they perform not better then the currently best Ada-optimizer Adam with optimal $\alpha$ and ultimately they converge in most cases to constant $\alpha$.\\
The use of a fixed learning rate $\alpha$ is in part due to the fact that it allows for precise mathematical analysis, guaranteeing or almost surely guaranteeing (for SGD) a lower bound on convergence rates (e.g.\ see Nesterov\cite{Nesterov:2018}, 1.2.3). The method (ELRA) to be proposed in this work is paradigm changing because it estimates in each step self-consistently a near optimal learning rate $\alpha$ for the next step from low-cost local knowledge of the function, thereby achieving a jump close to the next minimum along the gradient direction. In particular the learning rate approaches a problem-specific good scale exponentially fast. Therefore, we propose to name this class of optimizers \textbf{E}xponential \textbf{L}earning \textbf{R}ate \textbf{A}daption -- \textbf{ELRA}. Depending on the problem the adaption leads to continual substantial changes of $\alpha$.\\
Recent articles indicate that large variations of $\alpha$ might be very beneficial. In LongSteps2023\cite{grimmer2023longsteps}, it is for the first time mathematically proven that (periodically) varying step sizes lead to much better convergence rates, which is confirmed by our experimental results. In Truong2021\cite{Truong2021} it is shown that estimating the best $\alpha$ via backtracking using Armijo's, Nesterov\cite{Nesterov:2018}, 1.2.3) condition can lead to faster convergence then the Ada-family. However, each backtracking step needs a separate and expensive function value. Hence, backtracking more than once is seldom justified by the speed gained. Our algorithms do not suffer from this computational conundrum, as we provide two low-cost estimators for the best $\alpha$, thereby retaining the benefit of a good $\alpha$ without losing speed.\\
Note that such a strongly adaptive $\alpha$ completely eliminates the need for finding by hand a good constant $\alpha$ for a particular problem. Moreover, most modern training schemes rely on decreasing $\alpha$ over time to achieve faster convergence. The best timing is a priori unknown and often determined by an educated guess. We believe that a strongly adaptive $\alpha$ also needs no external timing. The third advantage is that our algorithms are invariant under orthogonal transformations of the arguments $x$, such as rotations, unlike the Ada-family. Such an invariance is preferable not only for geometric optimization (see VectorAdam2022\cite{vectoradam}) but also important near saddle points (see Results). 

\section*{Orthogonal gradients are optimal}
Let us briefly explain, how we estimate the best $\alpha$.
All gradient descent methods for minimizing a function $f$ boil down to the update scheme $x_t=x_{t-1}-\alpha((1{-}\beta)G_{t-1}+\beta M_{t-1})$ for the argument $x$ of $f$, where $G_{t-1}=\nabla f(x_{t-1})$ is the gradient at $x_{t-1}$, $M_{t-1}$ the momentum, $\beta$ the ratio between $G_{t-1}$ and $M_{t-1}$ (possibly zero). For the Ada-family, $\alpha$ is essentially constant while $G_{t-1}$ is not actually the gradient, but a component-wise modification, which is dynamically adapted. However, this leads to a dependency on the coordinate system and the speed of the algorithm depends heavily on the concrete representation of the data (see Figure \ref{fig:saddle_rot}). Moreover $\alpha$ has to be chosen with care, either using past results or initial try and error runs.\\
We provide a completely new approach which overcomes many of these problems. We can show that for the optimal learning rate $\alpha$, which leads locally to the smallest value $f(x_t)$, it holds that the new and previous gradients $G_t$ and $G_{t-1}$ are orthogonal to each other (see Methods, equation \eqref{eq:ortho}) or equivalently the cosine of the angle between the gradients $\cos_t:=\cos(\measuredangle(G_t,G_{t-1}))$ is zero. Moreover, we could even show that for $\cos_t<0$ the step size $\alpha$ should be decreased while for $\cos_t>0$ it should be increased. Figuratively speaking: If we see Zig-zag or anti-parallel steps we should decelerate, while for primarily parallel steps we should accelerate. This is computational much cheaper then Armijo's condition, as we need no extra gradient/function values.\\
We use two competing approaches to implement this idea to update $\alpha$. The first variant is $\alpha_t=\alpha_{t-1}(1+\cos_t/2)$. This formula for $\alpha$ is the core of our cosine-based optimizer \textbf{c2min}. The second variant is $\alpha_t=\alpha_{t-1}\big(1+\frac{\cos_t}{||G_{t-1}||/||G_t||-\cos_t}\big)$. This version requires no momentum ($\beta=0$). The update minimizes a parabola through $x_{t-1}$ and $x_t$ with slopes $-||G_{t-1}||$ and $-\langle G_{t-1},G_t\rangle$. This is the update formula for our parabola-based optimizer \textbf{p2min}. Note that for c2min, the updated step size $\alpha_t$ is bounded by $0.5\cdot\alpha_{t-1}$ and $1.5\cdot\alpha_{t-1}$, while it can be arbitrary between $-\infty$ and $+\infty$ for p2min. We prevent this potentially catastrophic behaviour by imposing bounds of the form $0<\alpha_t/\alpha_{t-1}<\gamma_{MAX}$, where $\gamma_{MAX}>1$ can be chosen at will, e.g.\ $\gamma_{MAX}\sim 10^6$. Moreover, we found that it is beneficial to set fixed bounds for $\alpha$. We use at the moment $10^{-8}<\alpha<10^6$. These are additional hyper-parameter, yet sufficient form almost all applications.\\
Note that an initial $\alpha_0>0$ still has to be chosen. However, the specific choice is only marginal, as both algorithms adapt $\alpha$ exponentially fast. We chose $\alpha_0$ very small (e.g.\ $\alpha_0\sim 10^{-5}$) to prevent initial instabilities (explosions of $f(x)$). This leads to a negligible fixed number of initial extra steps to increase $\alpha$ to the right magnitude (see Figure \ref{fig:mnist4s1p}).\\
Note that both approaches are by construction rotation invariant\footnote{Actually they are even invariant under orthogonal transformations.}, as they use only euclidean norms and cosines between vectors.  Moreover, their computation is relatively cheap (effort linear in dimension $n$, $O(n)$ time and space), as computing the norm and the cosine (or the scalar product) for vectors is relatively cheap.

\section*{Results}
We have a mathematical justification for our approach (see Methods, equation \eqref{eq:ortho}). Yet, giving estimates on guaranteed/expected convergence rates for our proposed optimizers is intractable using state of the arts methods (even if restricted to convex landscapes), due to the adaptive nature of the learning rate $\alpha$.  Thus we rely on experiments to show the usefulness of our optimizers.\\
All DNN experiments are executed for multiple starting points/initializations as gradient descent methods show partially chaotic behaviour\footnote{It appears to the authors that parts of the deep learning community are not fully aware of this fact.}, i.e.\ even small changes in initialization/batching can lead to drastically different optimization paths and minima.
However for cost reasons (limitations of an academical budget) we restrict ourselves to 5-25 different initializations per experiment and provide graphics which include the mean/median.
\subsection*{Mathematical 2D experiments}
As a proof of concept and to explore certain standard situations/problems in gradient descent, we first show results on 2-dimensional standard problems such as saddle points, bowls/parabolas and the Rosenbrock function.
\subsubsection*{Saddle points}
Saddle points (where $\nabla f(x)=0$ but $f(x)$ is not a local max/min) can pose problems in gradient descent methods, as the gradient becomes arbitrarily small near them, which might lead to catastrophic speed loss. Generically, for a suitable choice coordinates, these saddle points look locally like $x=(0,0)$ for $f(x_1,x_2)=x_1^2-x_2^2$  (see Math. Suppl., equation \eqref{eq:standardSaddle}). However, for a given data representation, it is more likely that the coordinates near a saddle are rotated!
\begin{figure}[ht]
\centering
\begin{subfigure}[b]{0.49\linewidth}
    \centering
    \includegraphics[width=\linewidth]{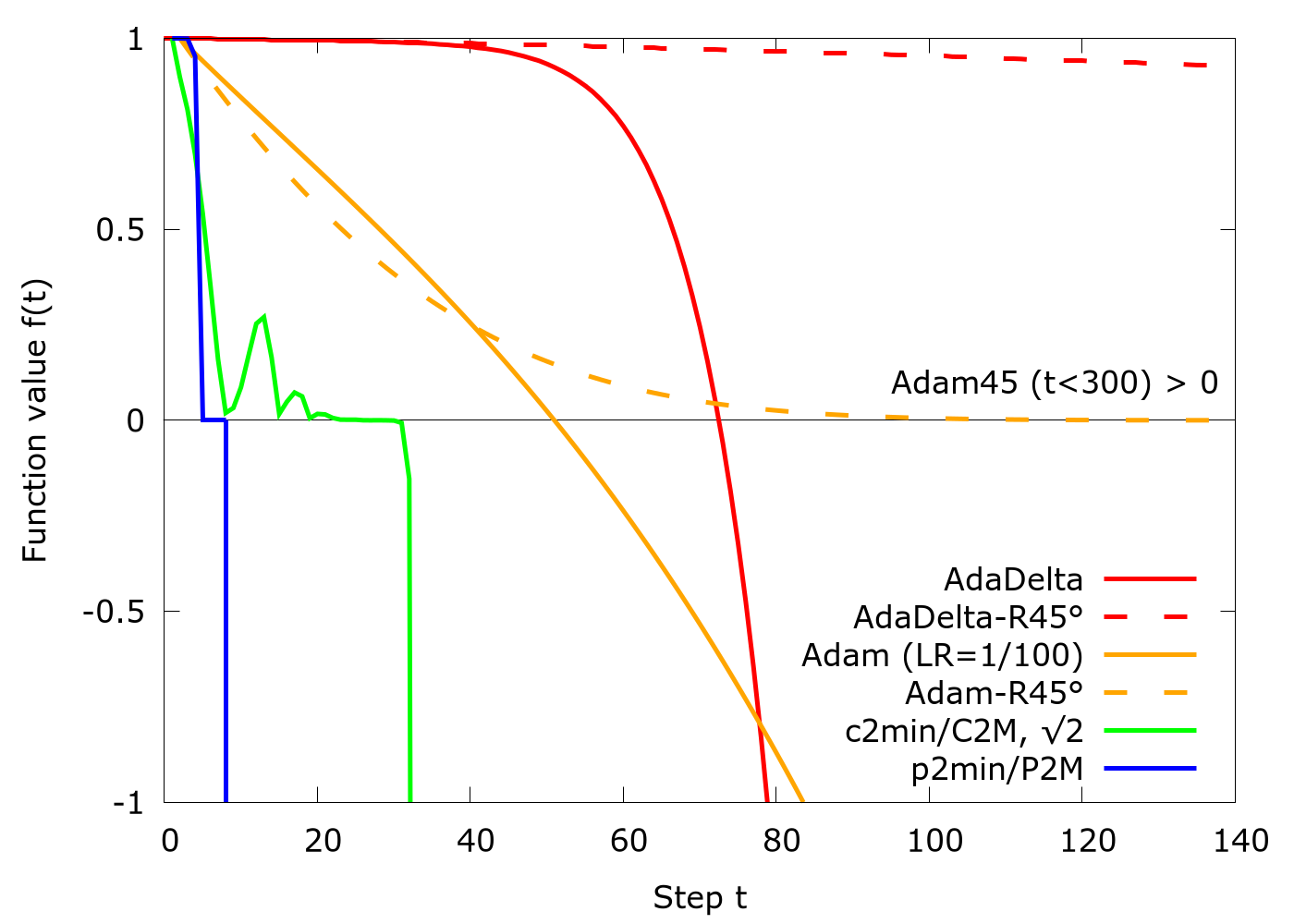} 
\caption{$f(x_1,x_2)=x_1^2-x_2^2$ over $steps  \;[t]$.}
\label{fig:saddle_rot}
\end{subfigure}
\hfill
\begin{subfigure}[b]{0.49\linewidth}
\centering
    \includegraphics[width=\linewidth]{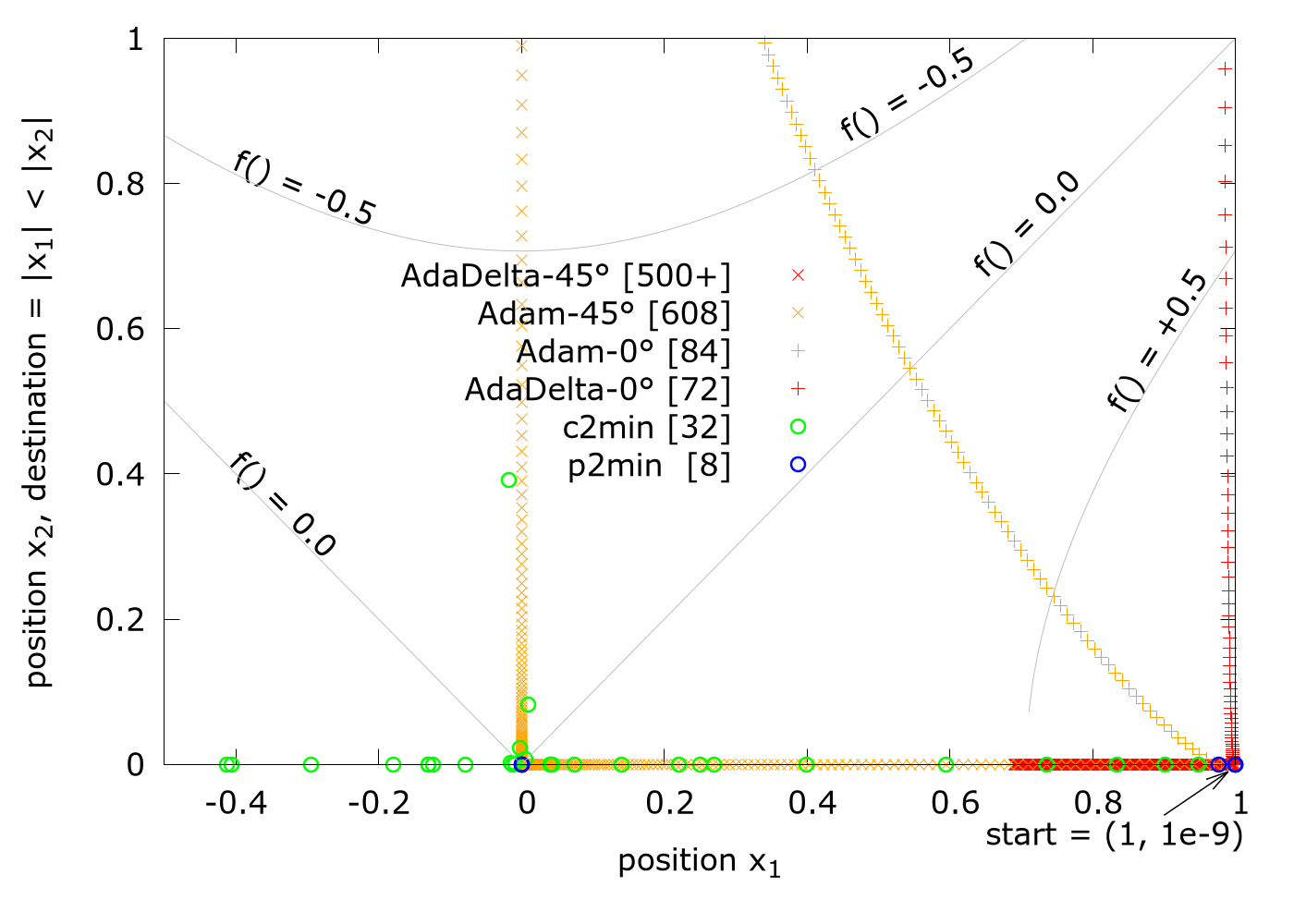} 
\caption{Paths in $x_1$-$x_2{-}$plane with $steps\;  [t]$ inside plot-region}
\label{fig:saddle_xy}
\end{subfigure}
\caption{Performance of optimizers near saddle and effect of $45^\circ$-rotation. p2min (blue) and c2min (green) are fastest (only 8 and 32 steps resp.\ to leave plot-region). AdaDelta (red) and Adam (orange) are slower in all cases, especially for rotated axes (dashed). Plot (b) illustrates the different paths of the optimizers ($\boldsymbol{\pmb{+}}\simeq0^\circ$, $\boldsymbol{\pmb{\times}}\simeq45^\circ$). Note that 4 out of the 8 steps of p2min (blue) are indistinguishable near origin $(0,0)$.}
\end{figure}\\
We looked at the performance of the optimizers AdaDelta, Adam (with $\alpha =0.01, \beta_1=0.9, \beta_2=0.999$), c2min and p2min near the standard saddle $f(x)=x_1^2-x_2^2$ starting at $x_0=(1,10^{-9})$ and the problem rotated by $45^\circ$. Figure \ref{fig:saddle_rot} shows the value of $f$ over steps $t$. The dashed lines belong to the rotated situation. The fastest are c2min and p2min and for each only one graph is visible (rotation invariance). AdaDelta and Adam are slower and suffer significantly from $45^\circ$-rotation, as it makes the component wise modification of the Ada-family completely useless. Figure \ref{fig:saddle_xy} illustrates the paths in the $x_1\text{-}x_2-$plane chosen by the different optimizers. One sees that c2min and p2min follow fast the gradient direction, while the Ada-family either try to avoid the saddle directly (unrotated situation) or follow slowly the gradient direction. This shows one drawback of conditioning individual axis weights within the Ada-family. It illustrates also that the different optimizers often find different local/global minima. Noteworthy: c2min (green) shows visible oscillations around the $x_2$-axis, which we use by design to decelerate.
\subsubsection*{Bowls and Rosenbrock}
As a second class of mathematical experiments, we considered higher dimensional parabolas (so called bowls), i.e.\ functions of the form $f(x)=\sum_i c_i\cdot x_i^2$, and the infamous Rosenbrock function $f(x_1,x_2)=(1{-}{x_1})^2+100 ({x_2{-}x_1^2})^2$. Bowls provide the simplest non-trivial functions for convex optimization, while the Rosenbrock function with its curved valley is a difficult standard optimization problem. Here, we used for Adam $\alpha=0.05, \beta_1=0.8, \beta_2=0.9$ and for RMSprop $\alpha=0.05$. The Tables \ref{tab:dataBowls}
 and \ref{tab:dataRosen} give the minimal number of steps $t$ needed for the different optimizers to reach a certain threshold for $f(x_t)$. One sees that for these examples (together with the saddle from above) p2min is by far the fastest and for Rosenbrock with bigger starting points, it is the only optimizer that produces any meaningful results. The bad  performance of c2min for Rosenbrock might be due to the constant momentum update. We hope to improve upon this result in the near future (see Future work).
\begin{table}[ht] 
\parbox{.49\textwidth}{\centering 
\begin{tabular}{|c||c|c|c|c|}\hline 
accuracy  & Adam & RMSprop & c2min & p2min\\\hline
$\varepsilon =10^{-1}$ & 128 & 142 & 28 & \textbf{9}\\\hline
$\varepsilon =10^{-6}$ & 184 & $\infty$ & 53 & \textbf{12}\\\hline
\end{tabular}
\caption{\label{tab:dataBowls} Steps $t$ to reach $f(x_t)<\varepsilon$ for bowl $f(x_1,x_2)=3x_1^2+24 x_2^2$ starting from $x_0=(-5.75,1.75)$}}
\hfill
\parbox{.49\textwidth}{\centering 
\begin{tabular}{|l||c|c|c|c|}\hline 
\hspace{.3cm} start point  & Adam & RMSprop & c2min & p2min\\\hline
$x_0=(-3,-2)$ & 208 & 176 & 1758 & \textbf{10}\\\hline
$x_0=(-11,121)$ & $>10^4$ & $>10^4$ & $>10^4$ & \textbf{300}\\\hline
\end{tabular}
\caption{\label{tab:dataRosen} Steps $t$ to reach $f(x_t)<1$ for Rosenbrock $f(x_1,x_2)=(1-{x_1})^2+100 ({x_2-x_1^2})^2$ starting from $x_0$}}
\end{table}
\pagebreak[4]
\subsection*{Neural networks}
\begin{wrapfigure}[20]{r}{0.7\linewidth}
    \vspace{-0.49cm}\includegraphics[width=0.99\linewidth]{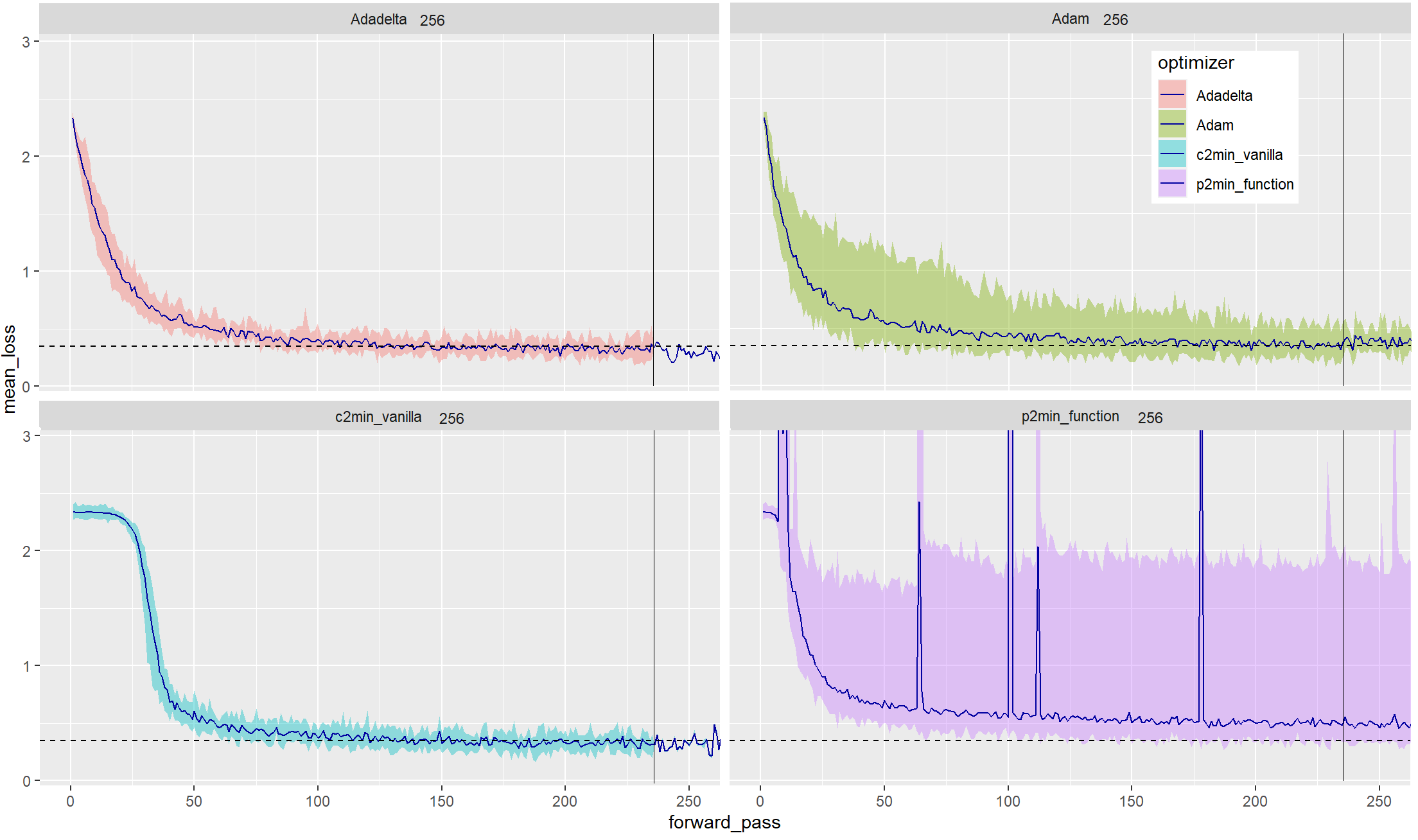}
\caption{MNIST mean \& min/max Test-loss (1st epoch $\cong$ black line), batch-size 256.}
\label{fig:mnist4s1p}
\end{wrapfigure}
\begin{figure}[ht]
\centering
\begin{subfigure}[b]{0.49\linewidth}
    \centering
    \includegraphics[width=\linewidth]{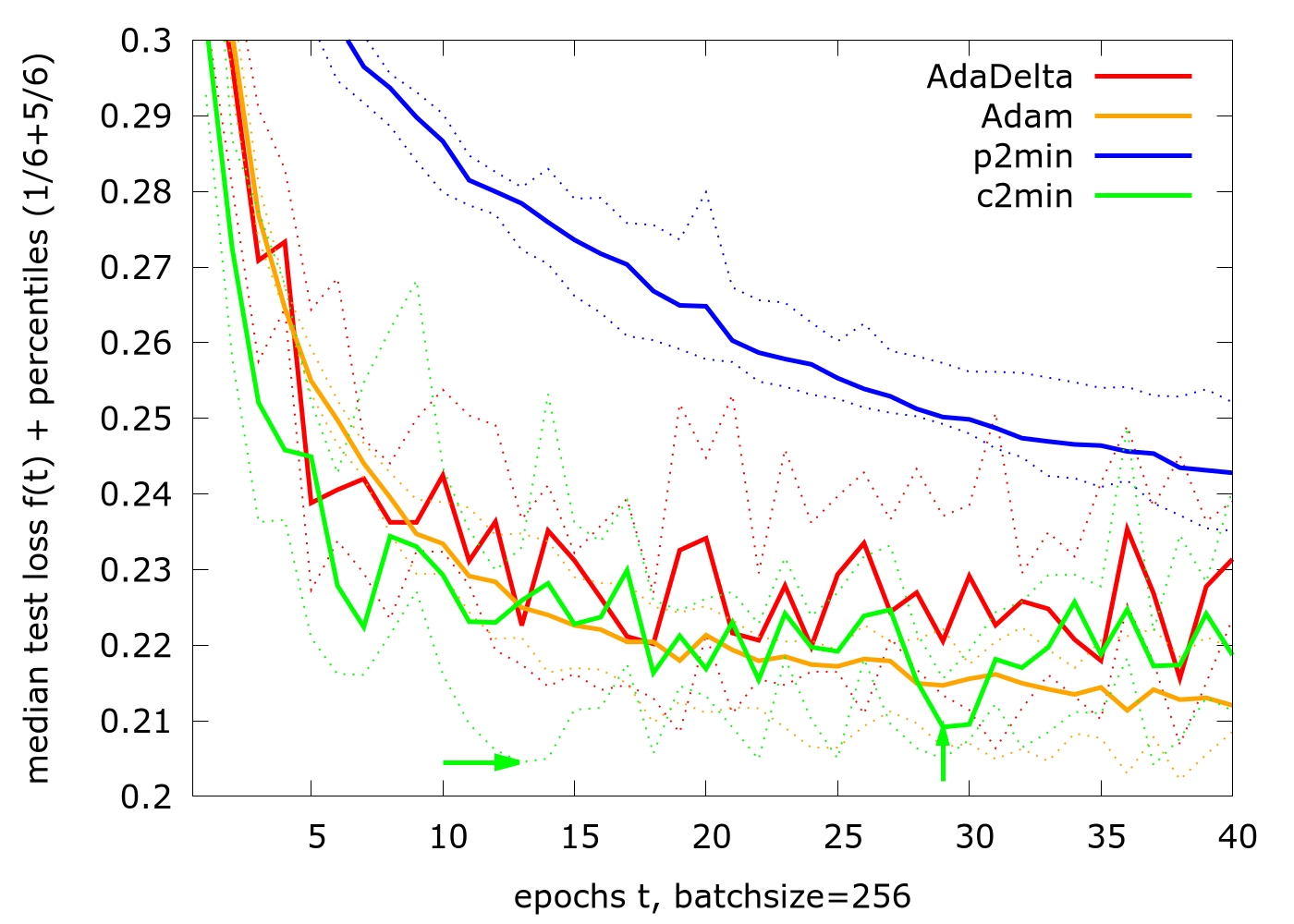} 
    \caption{Median, 1/6 \& 5/6 percentiles (dotted) of Test-Loss, Batch=256}
    \label{fig:test_perc}
\end{subfigure}
\hfill
\begin{subfigure}[b]{0.49\linewidth}
    \centering
    \includegraphics[width=\linewidth]{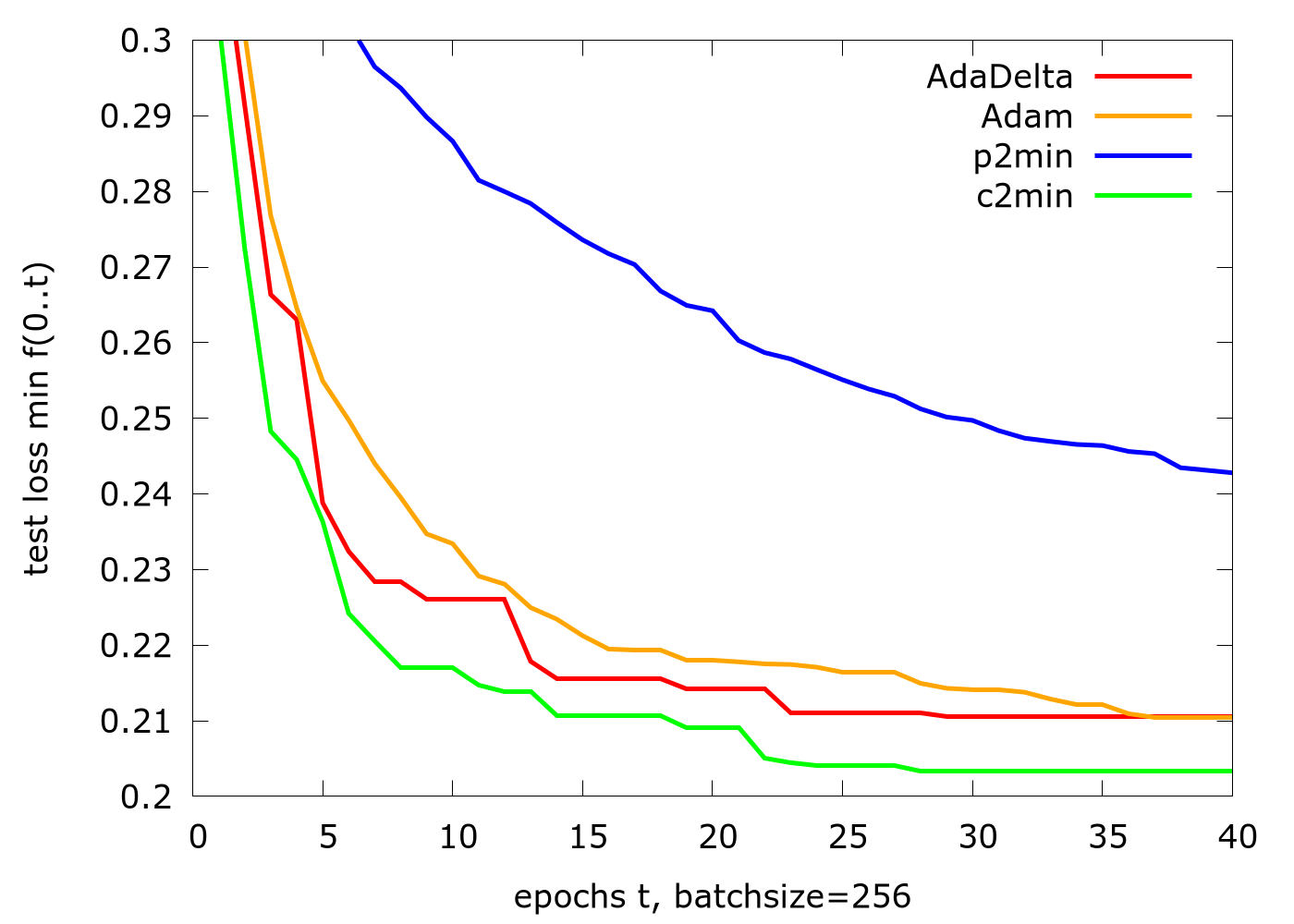} 
    \caption{Median of minimal Test-Loss up to $t$}
    \label{fig:best_loss_median}
\end{subfigure}
\caption{Test-Loss over 40 epochs (with 235 steps/epoch) using fine-tuned learning rate $\alpha=0.001$ for Adam.}
\end{figure}
\begin{figure}[ht]
\centering
\begin{subfigure}[b]{0.49\linewidth}
    \centering
    \includegraphics[width=\linewidth]{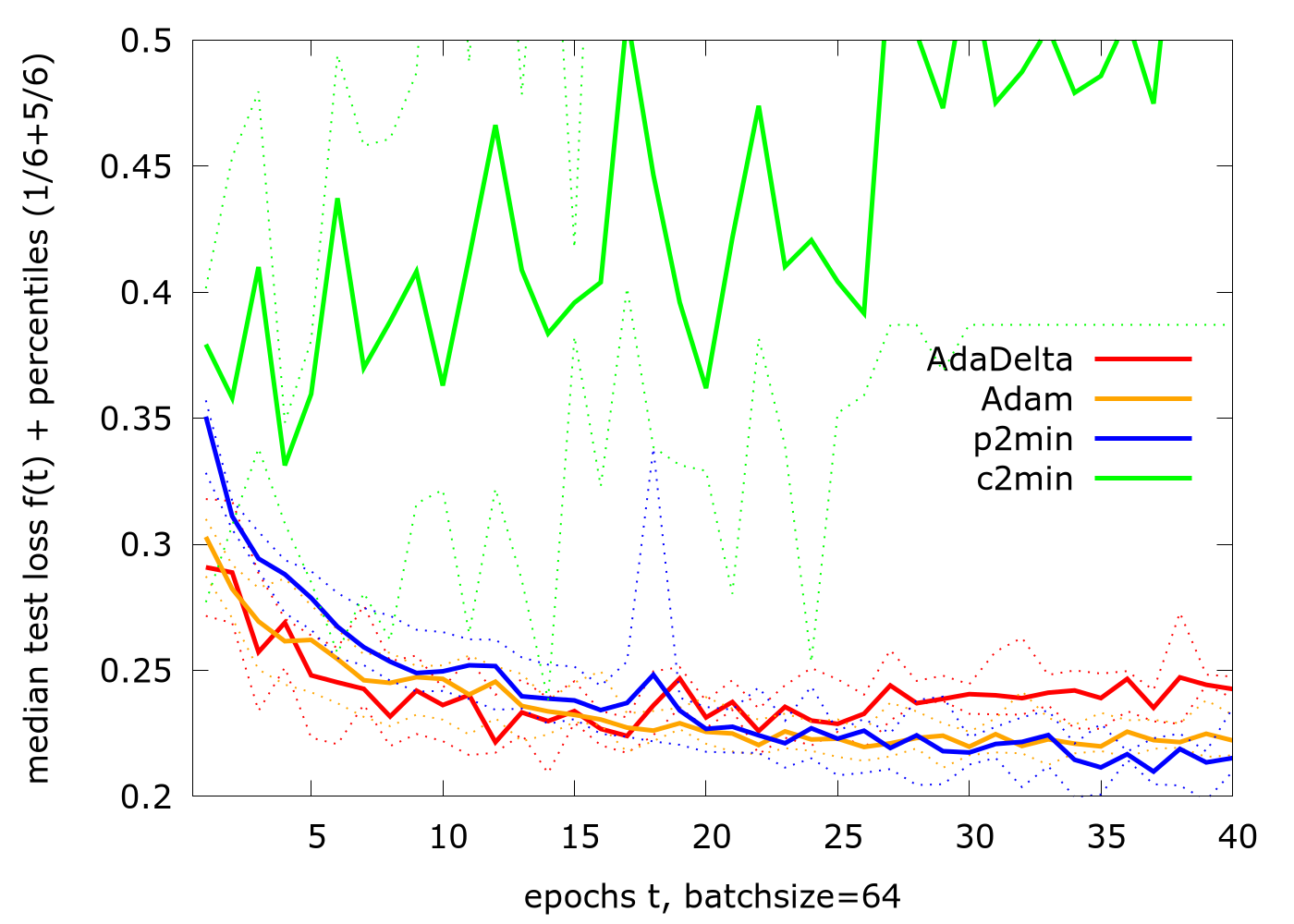} 
    \caption{Median, 1/6 \& 5/6 percentiles (dotted) of Test-Loss, Batch=64}
    \label{fig:test64_perc}
\end{subfigure}
\hfill
\begin{subfigure}[b]{0.49\linewidth}
    \centering
    \includegraphics[width=\linewidth]{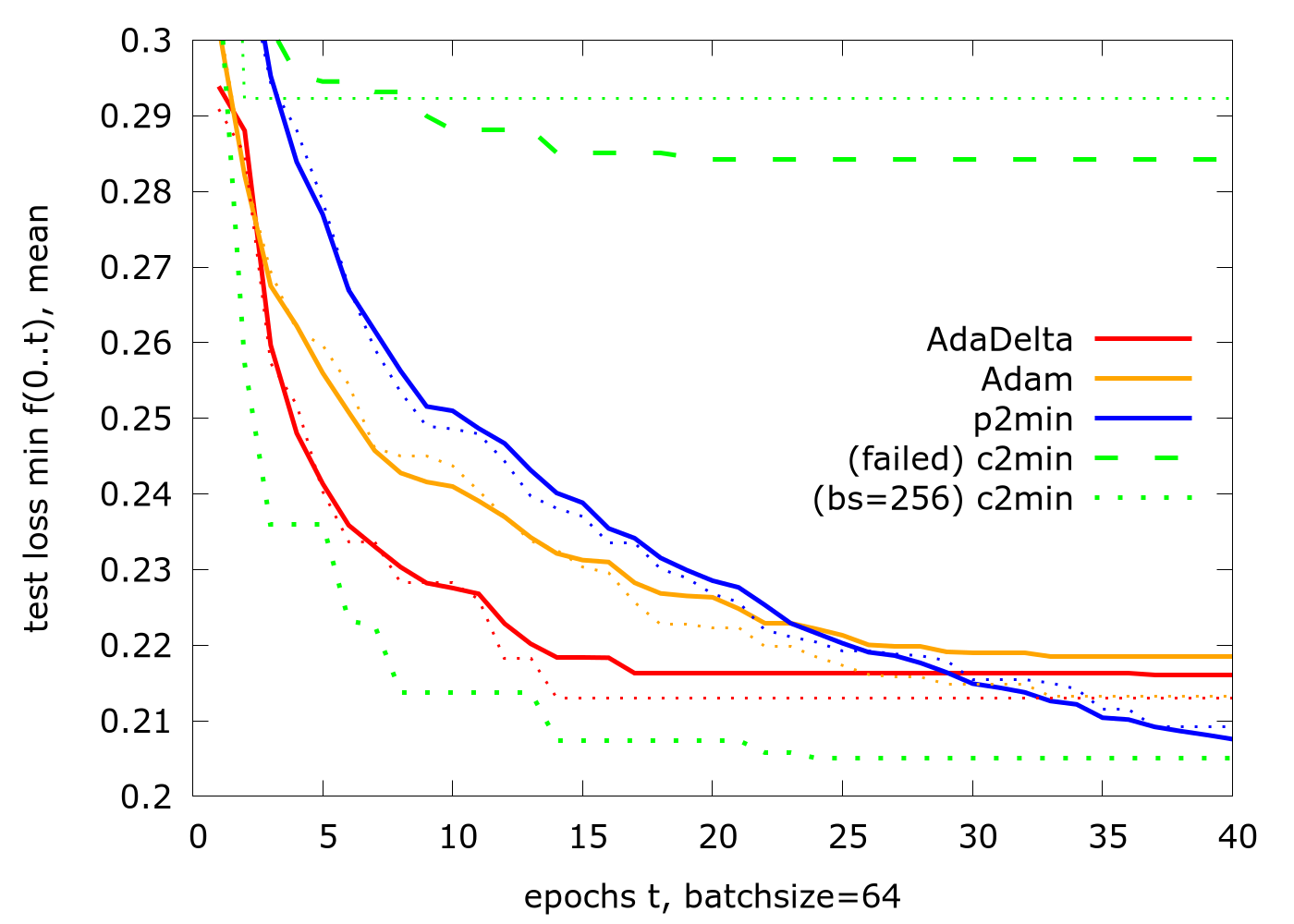} 
    \caption{Mean and median (dotted) of minimal Test-Loss up to $t$}
    \label{fig:best_loss64_median}
\end{subfigure}
\caption{Test-Loss over 40 epochs (with 470 steps/epoch) using fine-tuned learning rate $\alpha=0.001$ for Adam.}
\end{figure}
\begin{figure}[ht] \centering
\includegraphics[width=0.9\linewidth]{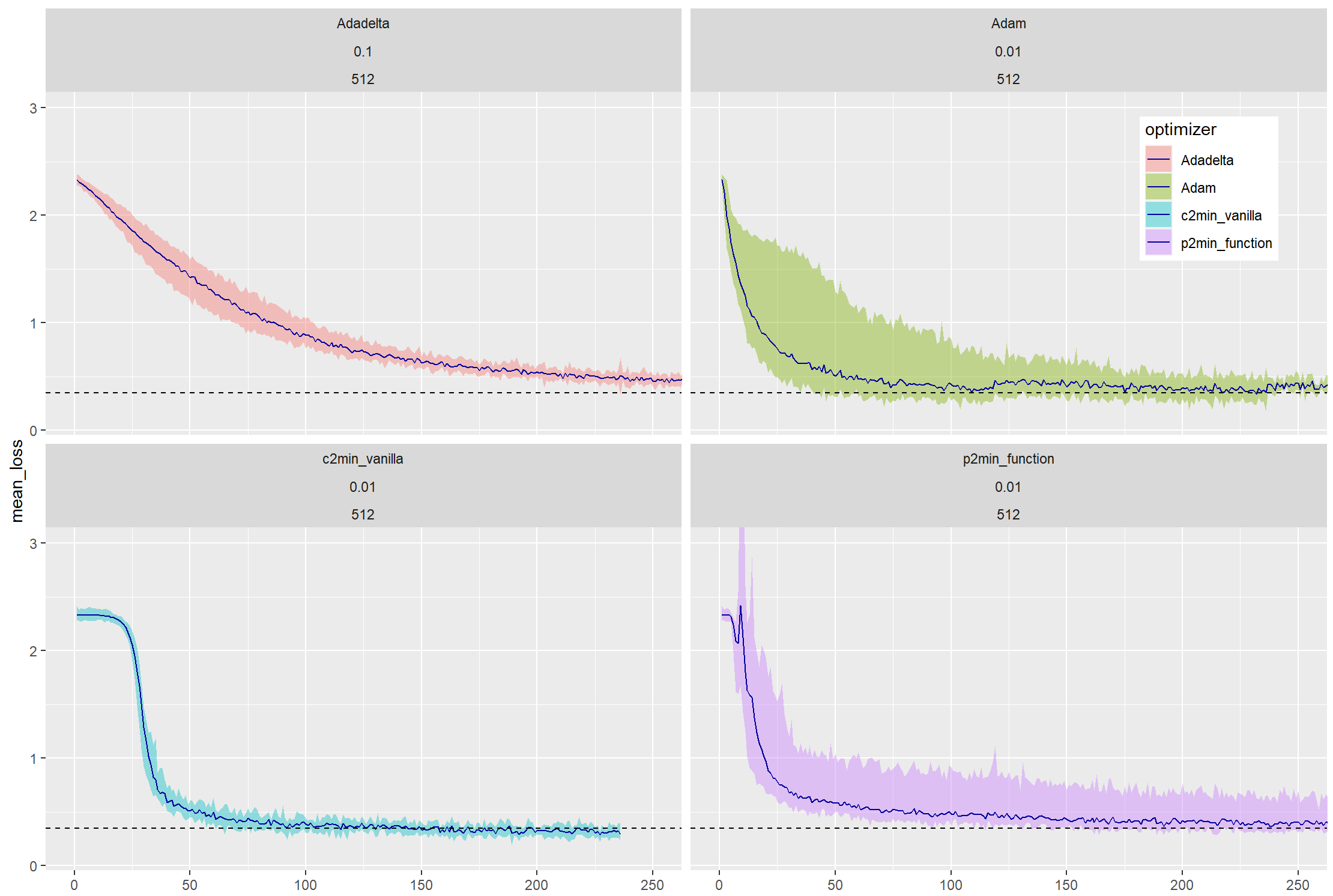}
\caption{MNIST mean \& min/max Test-loss (2 epochs, batch-size 512).}
\label{fig:mnist4s512}
\end{figure}

We conducted experiments on the MNIST data set for recognizing handwritten single digits from pictures consisting of 28x28 pixels. We used a simple fully connected network with 1 hidden layer (10 neurons) with ReLU-activation functions. This small design is to reduce computational costs. We also conducted few test with 2 hidden layers (16+16 neurons) giving similar results. We tested our two optimizers against the standard optimizers AdaDelta and Adam. First we performed 25 short runs over 1..4 epochs for each optimizer and batch-size to find an optimal global learning rate $\alpha$ for Adam (for short runs it is \linebreak[2] $\alpha=0.01$) and to test performance on different batch-sizes (256 (Fig. \ref{fig:mnist4s1p}), 512 (Fig.\ \ref{fig:mnist4s512}), 1024).
We see that c2min performs faster then the standard Ada-optimizers, while p2min's  performance is comparable, if one looks at the mean (dark blue curve). The plateau phase at the beginning of c2min comes from the very small initial learning rate $\alpha_0=10^{-5}$. This requires a fixed amount of steps to increase $\alpha$ to the right magnitude, due to exponential adaptation. The other optimizer p2min does not show this behaviour, as its adapts much faster. The spikes in the graph of p2min are due to dramatically bad estimations of $\alpha$, due to its aggressive adaptation of the learning rate. In these situations, we use a kind of soft restart (described below) correcting this behaviour.\\
Secondly, we conducted long run experiments spanning over 40 epochs, to determine best test loss results. Here, we found that $\alpha=0.001$ is the best learning rate for Adam. We conducted 8 training sessions with different (but same for every optimizer) initializations and training data shuffle after each epoch and batch size 256. This is not the best option for p2min. Therefore we conducted also training sessions with batch size 64. For batch size 256, we find at first glance a similar performance for AdaDelta, Adam and c2min (see Figure \ref{fig:test_perc}). Yet c2min reaches very good results fastest and finds the best results (green arrows). However, c2min enters later into a phase of oscillating test losses. Therefore, we plotted in Figure \ref{fig:best_loss_median} the minimal test losses from the first to the current epoch. This ignores potential later deterioration of the results.\\
Figures \ref{fig:test64_perc} and \ref{fig:best_loss64_median} show the results for micro-batches (size=64). Apparently, c2min fails here completely, while p2min now produces the best results after 30 epochs. Note that c2min with batch-size 256 still produces better and faster results.

\section*{Conclusion}
We presented a novel, simple, mainly self consistent, robust and fast optimizing method with linear dimensional scaling and rotational invariance, realized in two algorithms. Typical runs on mathematical standard problems and statistical tests on a neural network for the MNIST data set with several initializations showed better performance then the best state of the art optimizer Adam with hand-tuned optimal parameters(!).\\
We think that our algorithms still leave much room for improvement. Finding better control systems for alpha, momentum, soft restarts promise huge performance gains and increased universality (see Future works below). Meta-learning could also lead to further improvement. Moreover, sometimes c2min or p2min fail by decreasing $\alpha$ to much, essentially stopping the optimization midway. We believe to know the cause of these issues, yet to find a universal solution requires more time.\\
The authors thought about reasons why nobody tried steep and fast $\alpha$-variations before and see a couple of reasons: for small dimensions good solvers exist (often using matrix inversions, e.g.\ the Levenberg–Marquardt algorithm), mathematical optimizers strive for provability (which restricted until recently to constant $\alpha$: compare Nesterov\cite{Nesterov:2018} and LongSteps2023\cite{grimmer2023longsteps}) and previous conditions (Armijo) for updating $\alpha$ are often to expensive in high dimensions.

\bibliography{literature}

\begin{thebibliography}{1}
\urlstyle{rm}
\expandafter\ifx\csname url\endcsname\relax
  \def\url#1{\texttt{#1}}\fi
\expandafter\ifx\csname urlprefix\endcsname\relax\def\urlprefix{URL }\fi
\expandafter\ifx\csname doiprefix\endcsname\relax\def\doiprefix{DOI: }\fi
\providecommand{\bibinfo}[2]{#2}
\providecommand{\eprint}[2][]{\url{#2}}

\bibitem{coolmomentum}
\bibinfo{author}{Borysenko, O.} \& \bibinfo{author}{Byshkin, M.}
\newblock \bibinfo{journal}{\bibinfo{title}{Coolmomentum: a method for
  stochastic optimization by langevin dynamics with simulated annealing}}.
\newblock {\emph{\JournalTitle{Scientific Reports}}}
  \textbf{\bibinfo{volume}{11}}, \bibinfo{pages}{10705},
  \doiprefix\url{10.1038/s41598-021-90144-3} (\bibinfo{year}{2021}).

\bibitem{psocitation}
\bibinfo{author}{Ab~Wahab, M.~N.}, \bibinfo{author}{Nefti-Meziani, S.} \&
  \bibinfo{author}{Atyabi, A.}
\newblock \bibinfo{journal}{\bibinfo{title}{A comprehensive review of swarm
  optimization algorithms}}.
\newblock {\emph{\JournalTitle{PloS one}}} \textbf{\bibinfo{volume}{10}},
  \bibinfo{pages}{e0122827} (\bibinfo{year}{2015}).

\bibitem{mishchenko2023prodigy}
\bibinfo{author}{Mishchenko, K.} \& \bibinfo{author}{Defazio, A.}
\newblock \bibinfo{title}{Prodigy: An expeditiously adaptive parameter-free
  learner} (\bibinfo{year}{2023}).
\newblock \eprint{2306.06101}.

\bibitem{ivgi2023dog}
\bibinfo{author}{Ivgi, M.}, \bibinfo{author}{Hinder, O.} \&
  \bibinfo{author}{Carmon, Y.}
\newblock \bibinfo{title}{Dog is sgd's best friend: A parameter-free dynamic
  step size schedule} (\bibinfo{year}{2023}).
\newblock \eprint{2302.12022}.

\bibitem{Nesterov:2018}
\bibinfo{author}{Nesterov, Y.}
\newblock \emph{\bibinfo{title}{Lectures on Convex Optimization}}
  (\bibinfo{publisher}{Springer Publishing Company, Incorporated},
  \bibinfo{year}{2018}), \bibinfo{edition}{2nd} edn.

\bibitem{grimmer2023longsteps}
\bibinfo{author}{Grimmer, B.}
\newblock \bibinfo{title}{Provably faster gradient descent via long steps}
  (\bibinfo{year}{2023}).
\newblock \eprint{2307.06324}.

\bibitem{Truong2021}
\bibinfo{author}{Truong, T.~T.} \& \bibinfo{author}{Nguyen, H.-T.}
\newblock \bibinfo{journal}{\bibinfo{title}{Backtracking gradient descent
  method and some applications in large scale optimisation. part 2: Algorithms
  and experiments}}.
\newblock {\emph{\JournalTitle{Applied Mathematics \& Optimization}}}
  \textbf{\bibinfo{volume}{84}}, \bibinfo{pages}{2557--2586},
  \doiprefix\url{10.1007/s00245-020-09718-8} (\bibinfo{year}{2021}).

\bibitem{vectoradam}
\bibinfo{author}{Ling, S.}, \bibinfo{author}{Sharp, N.} \&
  \bibinfo{author}{Jacobson, A.}
\newblock \bibinfo{title}{Vectoradam for rotation equivariant geometry
  optimization}, \doiprefix\url{10.48550/ARXIV.2205.13599}
  (\bibinfo{year}{2022}).

\bibitem{Milnor:1965}
\bibinfo{author}{Milnor, J.}
\newblock \emph{\bibinfo{title}{Lectures on the H-Cobordism Theorem}}
  (\bibinfo{publisher}{Princeton University Press},
  \bibinfo{address}{Princeton}, \bibinfo{year}{1965}).

\end{thebibliography}



\newpage
\section*{Appendix}
\section*{Methods}

\subsection*{Fundamental Concept of using Cosine (Orthogonal gradients are optimal)}\label{sec:cosine}
The main idea of our algorithms is to use the cosine between the new gradient $ G_t$ and the previous gradient $G_{t-1}$ to decide how the learning rate $\alpha$ should be adapted. First we prove why this is mathematically reasonable: We want to find a local minimizer $x_t$ to the differentiable\footnote{See Mathematical supplements, equation \eqref{eq:NoiseSmooth}, why even for non-differentiable activation functions, such as ReLU, one can assume that $f$ is smooth.} function $f$ near a non-critical point $x_{t-1}$ in the negative gradient direction $-G_{t-1}$. We consider $h(\alpha):=f(x_{t-1}-\alpha\cdot G_{t-1})=f(x_t)$, which is the value of $f$ at the next point $x_t$, depending on the learning rate $\alpha$. Differentiating $h$ with respect to $\alpha$ yields by the chain rule:
\begin{equation}\label{eq:ortho}
    h'(\alpha)=\big\langle \nabla f(x_t),-G_{t-1}\big\rangle=-\big\langle G_t,G_{t-1}\big\rangle=-\cos\big(\measuredangle(G_t\text{\large,}G_{t-1})\big)\cdot||G_t||\cdot||G_{t-1}||.
\end{equation}
Note that $h'(0)=-||G_{t-1}||^2$ is negative. This means that $h$, and hence $f$, decreases for small $\alpha$. This continues to hold right up to the first critical point $\alpha_{min}>0$ of $h$, where we have $h'(\alpha_{min})=0\,\Leftrightarrow \, \cos_t:=\cos(\measuredangle(G_t,G_{t-1}))=0$. If $\alpha_{min}$ is a non-degenerate critical point of $h$, then $h$ has necessarily a local minimum at $\alpha_{min}$ and thus also $f$ in the direction of $-G_{t-1}$. This gives us the desired conclusion: For $\alpha\in [0,\alpha_{max}]$ for some positive constant $\alpha_{max}$ (the second positive critical point of $h$), it holds that if $\cos_t=0$ then $\alpha$ was optimal, if $\cos_t>0$ then an increased $\alpha$ would have given better results, and $\cos_t<0$ a smaller $\alpha$ would have been better. As $\alpha_{min}$ depends continuously on $x_{t-1}$, we can expect that the optimal $\alpha_t$ for $x_t$ does not vary too much from the optimal $\alpha_{t-1}$ for $x_{t-1}$. This justifies the use of $\cos_t$ as an oracle for the next $\alpha_t$.\\
There are infinitely many ways to use this result to update $\alpha$, which can all be written in the form $\alpha_t=\alpha_{t-1}(1+\cos_t\cdot g)$, where $g$ can be any function, provided $g>0$. We tried four possibilities. The first is $\alpha_t=\alpha_{t-1}\cdot a^{\cos_t}$ for a constant $a>1$. The choice of $a$ sets the maximal/minimal factor, by which $\alpha$ is changed, here $\alpha_t=\alpha_{t-1}/a$ for $\cos_t=-1$ and $\alpha_t=\alpha_{t-1}\cdot a$ for $\cos_t=1$. We propose $a=\sqrt{2}$ leading to $\alpha_t=\alpha_{t-1}\cdot\sqrt{2}^{\cos_t}$ for a mild exponential growth. The linearization of $a^{\cos_t}$ at zero gives the second approach $\alpha_t=\alpha_{t-1}\cdot(1+\cos_t\cdot\ln a)$. Explicitly, we choose $a=\sqrt{e}$, to get $\alpha_t=\alpha_{t-1}(1+0.5\cdot\cos_{t-1})$, which has a similar performance, but involves easier computations and has an asymmetric update behaviour, as here $\alpha_t=\alpha_{t-1}/2$ for $\cos_t=-1$ and $\alpha_t=\alpha_{t-1}\cdot 3/2$ for $\cos_t=1$, i.e.\ we reduce $\alpha$ faster and increase $\alpha$ slower, which leads to a more conservative and hence stable behaviour. Another update-scheme for $\alpha$ is:
\begin{equation}\label{eq:parabel1}\begin{aligned}
    \alpha_t=\alpha_{t-1}\cdot\frac{||G_{t-1}||^2}{||G_{t-1}||^2-\langle G_{t-1},G_{t}\rangle}&=\alpha_{t-1}\cdot\left(1+\frac{\langle G_t,G_{t-1}\rangle}{||G_{t-1}||^2-\langle G_{t-1},G_{t}\rangle}\right)\\
    &=\alpha_{t-1}\cdot\left(1+\frac{\cos_t\cdot||G_{t-1}||\cdot||G_t||}{||G_{t-1}||^2-\cos_t\cdot ||G_{t-1}||\cdot||G_{t}||}\right)\\
    &=\alpha_{t-1}\cdot\left(1+\frac{\cos_t}{||G_{t-1}||\cdot||G_t||^{-1}-\cos_t}\right)\quad\qquad=\alpha_{t-1}\cdot h_t.
    \end{aligned}
\end{equation}
Here, $\alpha_t$ is chosen such that $(x_{t-1}{-}\alpha_t\cdot G_{t-1})$ is the minimizer of a parabola through $x_{t-1}$ and $x_t$ with slopes $-||G_{t-1}||$ and $-\langle G_{t-1},G_t\rangle$ in $x_{t-1}$  and $x_{t}$ respectively. Again $\alpha_t\lesseqqgtr\alpha_{t-1}$ for $\cos_t\lesseqqgtr 0$ (provided $||G_{t-1}||/||G_t||-\cos_t>0$). Using the vertex of a parabola determined by the slope $-||G_{t-1}||$ at $x_{t-1}$ and the function values $f(x_{t-1})$ and $f(x_t)$ leads to the last version:
\begin{equation}\label{eq:parabel2}
    \alpha_t=\alpha_{t-1}\cdot\frac{\alpha_{t-1}\cdot||G_{t-1}||^2}{2(f(x_t)-f(x_{t-1})+\alpha_{t-1}\cdot||G_{t-1}||^2)}=\alpha_{t-1}\cdot\hat{h}_{t}
\end{equation}
This last formula does not share the cosine property of the other three approaches, but behaves similar to the third. Note that \eqref{eq:parabel1} and \eqref{eq:parabel2} can in principle change $\alpha$ by an arbitrary factor $-\infty\leq h_t\leq \infty$. In practice, we restrict $h_t$, in particular for almost linear cases ($h_t=\infty$) and wrong sided vertices for upside down parabolas ($h_t<0$), by requiring $1/h_t\geq 1/\gamma_{MAX}$ and setting $h_t=\gamma_{MAX}$ otherwise\footnote{The use of the reciprocal value allows us to test for $h_t<0$ and $h_t>\gamma_{max}$ simultaneously.}.

\subsection*{Momentum Oscillation reduction (gradient pair averaging)} 
Momentum is a known strategy to solve/improve two issues: flat regions and damping of high freq. oscillations (self-produced noise).
Many unfriendly (often non-convex) landscapes (e.g. Rosenbrock, $|x|$) but also multi-dimensional bowls ($\sum a_i x_i^2$) show such oscillations and resulting reduced convergence rates. The Ada-family uses simple decaying average of gradients for momentum $M$. We suggest updating the momentum with sums of pairs of gradients $(G_t+G_{t-1})/2$, which leads for \textbf{c2min} to the update-scheme $M_t=\beta_1\cdot M_{t-1}+(1-\beta_1)\cdot(G_t+G_{t-1})/2$, where we use $\beta_1 =0.8$. It is important to observe, that we use a different parameter $\beta_2=0.7$ for the actual step $x_t-x_{t-1}=-\alpha_t((1-\beta_2)G_{t-1}+\beta_2M_{t-1})$. This choice is at the moment purely phenomenological (the best choice for us) and leaves room for future improvement. Therefore, the momentum is not free of hyper-parameters. We have some ideas, how to adapt $\beta_1,\beta_2$ dynamically (see Future work), but have not included them here for simplicity.
 
 \subsection*{Soft restart and variable maximal growth rate}
 The update -scheme for $\alpha$ described above in \textbf{p2min} is quite aggressive and would often lead to numerical instability. To prevent these, we use a form of retracing/soft restart, if the new value $f(x_t)$ increases too much\footnote{We need to allow some increment, as we are dealing with stochastic gradients.}. Then we retrace back to the previous $x_{t-2}$, calculate $\alpha_t$ using formula \eqref{eq:parabel2} (as it guarantees reducing $\alpha$ by a factor of at least 0.5 in this case). The next argument $x_t$ is then calculated as
 \[x_t=x_{t-2}-\alpha_tG_{t-2} = x_{t-1}+\alpha_{t-1}G_{t-2}-\alpha_tG_{t-2}.\]
 The second formula shows, that we do not have to store $x_{t-2}$ for the retracing, thus saving memory.\\
 The question, when to retrace, is quite a delicate one and leaves room for much improvement. At the moment, it would suffice for neural networks to retrace if $f(x_t)>25\cdot f(x_{best})$, where $f(x_{best})$ is the best function value seen thus far. For neural networks, $f(x_{best})$ should be the best average training loss over a whole epoch. For the more mathematical experiments, such as Rosenbrock, we actually use a more involved condition of the form $f(x_t)<f(x_{best})/D$ including a damper variable $D$, which increases if $f(x_t)>f(x_{t-1})$ and decreases otherwise.\\
 The maximal growth rate $\gamma_{max}$ for $\alpha$ is another aspect with much room for future improvement and research. For neural networks, it probably suffices to set $\gamma_{max}=10$, while for more complicated landscapes, we actually use $\gamma_{max}=10^6/(1+d)$, where $d$ is another damper, ranging from $0$ to $10^6$. We increase $d$, whenever we have a soft restart and decrease it otherwise.

\section*{Future work}
\begin{itemize}
    \item $\alpha=\alpha\cdot(1+\cos\cdot g(x))$ is the general update scheme for $\alpha$ obtained from our idea of orthogonal gradients \eqref{eq:ortho}. Here, $g$ can be any function with $g(x)>0$. What is the best $g$? Different answers for different problems? E.g. $(1+\cos /c)$ gives for $c>0.5$ faster and for $c<0.5$ slower adaptions of $\alpha$ compared to the currently used $c=0.5$ in c2min.
    \item short initial phase to find faster ideal initial $\alpha_0$ (eliminates need to chose $\alpha_0$ very small)
    \item Problem specific fine tuning (selected hyper-parameters) is possible and could give further improvement.
    \begin{itemize}
        \item fixed bounds for $\alpha$ (i.e.\ $10^{-7}<\alpha<10^{-1}$) based on statistics gathered during current run. Could speed up p2min significantly (fewer soft restarts, shorter time to recover from restart. 
        \item dynamic adaption of momentum parameters $\beta_1, \beta_2$ for c2min
    \end{itemize}    
    \item Other fields of optimization (e.g. electronic structure) should give this a try.
    \item Possible landscape characterization as a side-result.
 \end{itemize}
 \begin{figure}[ht] \centering
\includegraphics[width=0.5\linewidth]{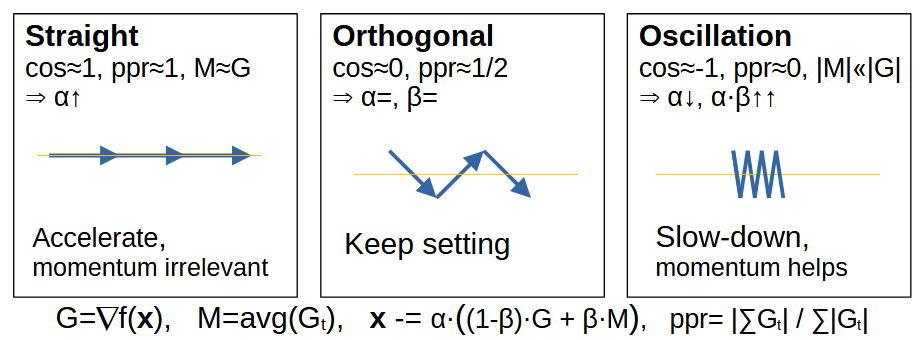}
\caption{General situations during optimization. Controlling $\alpha$ is most important, where we consider cosine between successive gradients as dominant input. If path-progress-ratio $ppr=|\sum G_t|/\sum |G_t|\approx 0$, momentum is beneficial.}
\label{fig:phases3}
\end{figure}

\section*{Mathematical supplements}
\subsection*{Extremal points sit inside quadratic surrounding} In principle, critical points $x_0$, such as local/global minima and saddle points, can be degenerate, i.e.\ the Hessian at $x_0$ can have 0 as an eigenvalue. However, functions with all critical points non-degenerate, so called Morse functions, are the generic situation, meaning that they form an open and dense subset\cite{Milnor:1965} within $C^2(\mathbb{R}^n$). So figuratively speaking, "almost all" two times continuously differentiable functions have only non-degenerate critical points. For these functions $f$, we find then by Taylor expansion, that they are locally dominated by their Hessian, i.e.\ they behave locally around critical points like quadratic functions:
\begin{equation}\label{eq:standardSaddle}
    f(x)=\sum_{i=1}^n c_i\cdot x_i^2, \qquad c_i\in\{+1,-1\}.
\end{equation} 

\subsection*{Random noise convolution removes discontinuities in Gradients}
In some applications, the function $f$, which we want to optimize, is not differentiable, such as $f(x)=||x||$ or $f(x)=\max\{x,0\}$. Then, the gradient is not everywhere defined and most optimization methods suffer. However, if the data contains some random noise, i.e.\ the function $f$ is slightly blurred, then we can expect differentiability. Indeed, the effect of random noise can be thought of as convoluting $f$ with a probability density function $\phi$, such as the density of the normal distribution
$\phi(x)=exp(-x^2/2\sigma^2)/(\sigma\sqrt{\pi})$ 
(if the blurring can be arbitrarily large) or a density with finite support, if the blurring is limited. Now, if $\phi$ is continuously differentiable and $f$ integrable or locally integrable (for finite support), then it is a well known fact that the convolution $f\ast\phi$ is also differentiable with differential
\begin{equation}\label{eq:NoiseSmooth}
    \partial_{x_i}(f\ast\phi)(x)=\partial_{x_i}\int_{\mathbb{R}^n} f(t)\cdot\phi(x-t)\text{d}t=\int_{\mathbb{R}^n}f(t)\cdot\partial_{x_i}\phi(x-t)\text{d}t=\left(f\ast\partial_{x_i}\phi\right)(x).
\end{equation}
Especially DNN learning should be affected by noise from the input and from batching, resulting in smooth landscapes.

\section*{Acknowledgements}
We are grateful to the Technical University of Applied Sciences Wildau for giving us the possibility to do applied research.

\section*{Author contributions statement}

A.K. brought up the core idea, S.K. and A.K. implemented c2min, A.F. worked out the mathematical details and developed c2min, F.R. created the automated test suite and contributed with DNN insight knowledge. All authors reviewed the manuscript. 


\section*{Additional data}
\begin{table}[ht] \centering
\begin{tabular}{|l|l|l|l|l|l|l|l|l|l|l|} \hline
Method & VecCnt & O(dim) & NoLR & Non-Conv & Speed & Stable & Rot-Inv. & ScaleX-Inv. & Noise & SadMf\\ \hline
sGD & 2 & + & - & +/- & - & -- & + & - & ++ & - \\ \hline
LMA & $2+2\cdot d$ & - & + & - & + & -- & + & ++ & - & (?) \\ \hline
\hline
RMSProp & 4 & + & - & + & + & + & - & - & = & -  \\ \hline
AdaMax & 4 & + & - & = & (+) & + & - & - & = & -  \\ \hline
AdaDelta & 4 & + & + & + & + & + & - & + & = & -  \\ \hline
Adam & 4 & + & - & + & + & (+) & - & - & = & -  \\ \hline
SFN & $2+2\cdot d$ & - & ? & + & + & ++ & + & + & ? & +  \\ \hline
VecAdam & $\geq 4$ & - & - & + & + & ++ & + & - & = & -  \\ \hline
AdaSmoo. & 5 & + & + & + & + & + & - & (+) & +? & ?  \\ \hline
Tadam & 4 & + & - & + & + & + & - & - & + & -  \\ \hline
AggMo & $\geq 5$ & + & - & + & + & + & - & - & ? & - \\ \hline
\hline
c2min & 4 & + & + & + & ++ & ++ & + & ++ & +? & + \\ \hline
p2min & 3 & + & + & + & +++ & ++ & + & ++ & +? & + \\ \hline
\end{tabular}
\caption{\label{tab:algos}Rough comparison of optimization algorithms. Only our methods combine the following properties (left to right column), number of vectors, linear dimension scaling, free of learning rate, non-convex handling, convergence speed, broad stability, rotational invariance, independent of parameter scaling, noise robustness, saddle points w/o momentum capability.}
\end{table}
\pagebreak[3]

\begin{figure}[ht] \centering
\includegraphics[width=0.4\linewidth]{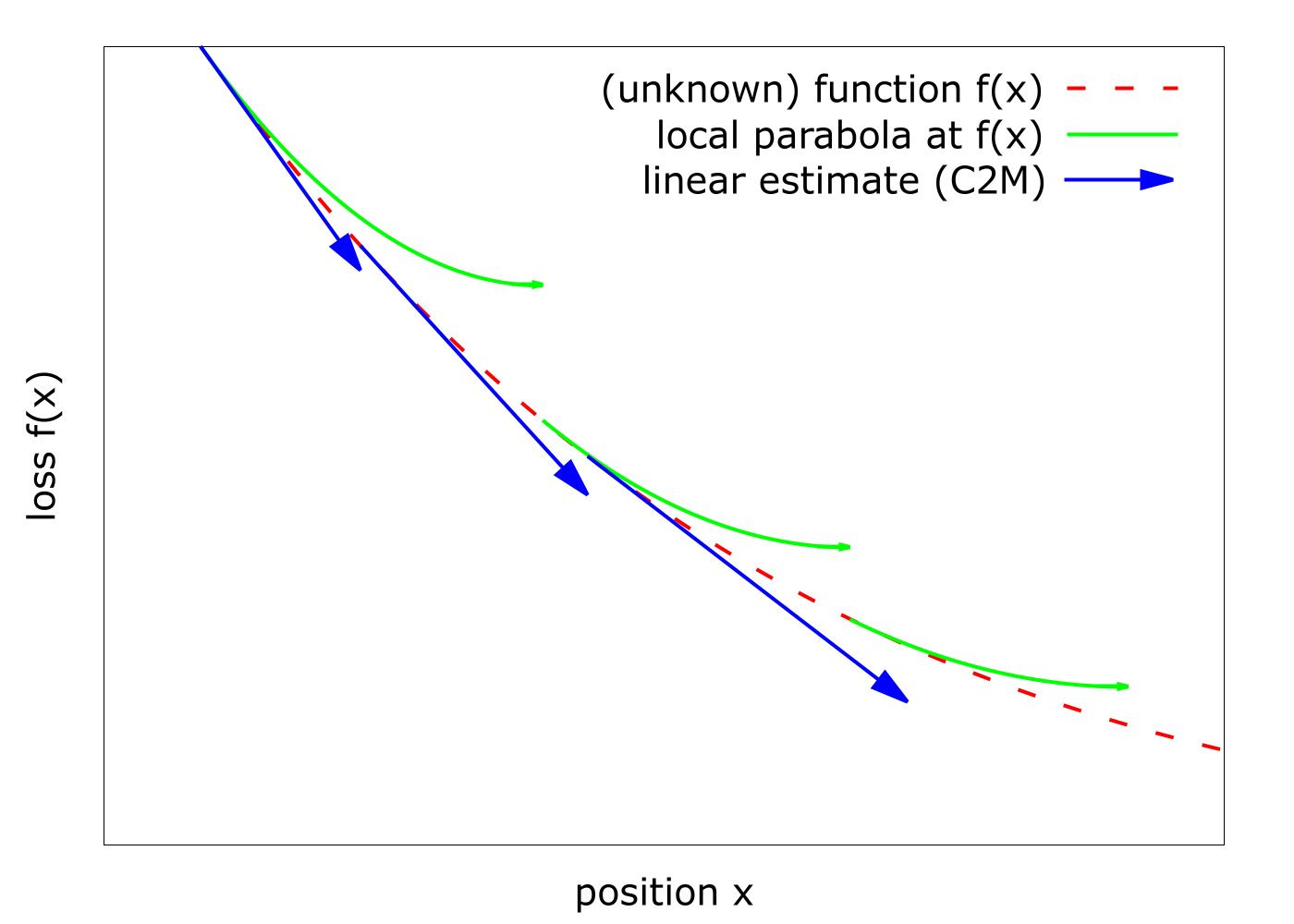}
\caption{Schematic view of the first 3decent steps for P2M (green) and C2M (blue). Both solver variants follow the negative gradient with different $\alpha$ scaling. P2M follows the local estimated parabola until its vertex. C2M is scaling up $\alpha$ as long the gradients-pairs are mainly parallel.}
\label{fig:pathplot}
\end{figure}
\end{document}